\documentstyle{article} 
\newcommand{\lea}{\stackrel{+}{<}}
\newcommand{\gea}{\stackrel{+}{>}}
\newcommand{\eqa}{\stackrel{+}{=}}

\newcommand{\G}{\mbox{\rm G}}

\newtheorem{definition}{Definition}

\title{The Generalized Universal Law of Generalization}

\author{
Nick Chater\thanks{Partially supported by European Commission grant
RTN-HPRN-CT-1999-00065.
Address: Department of Psychology
University of Warwick
Coventry, CV4 7AL, UK. Email:
nick.chater@warwick.ac.uk}
\\University of Warwick
\and
Paul M.B. Vit\'{a}nyi\thanks{Partially
supported by the EU through NeuroColt II Working Group and the QAIP Project.
Address: Centrum voor Wiskunde en Informatica,
Kruislaan 413, 1098 SJ Amsterdam, The Netherlands. Email: paulv@cwi.nl}\\
CWI and Universiteit van Amsterdam}

\begin{document}

\maketitle

\begin{abstract}
It has been argued by Shepard
that there is a robust psychological law that
relates the distance between a pair of items in psychological space and the
probability that they will be confused with each other. Specifically, the
probability of confusion is a negative exponential function of the distance
between the pair of items. In experimental contexts, distance is typically
defined in terms of a multidimensional Euclidean space---but this assumption
seems unlikely to hold for complex stimuli. We show that, nonetheless, the
Universal Law of Generalization can be derived in the
more complex setting of arbitrary stimuli, using
a much more universal measure of distance. This universal
distance is defined as the length of the shortest
program that transforms the representations of the two items
of interest into one another: the algorithmic information distance.
It is universal in the sense that it minorizes every computable distance:
it is the smallest computable distance.
We show that
the universal law of generalization
holds with probability going to one---provided the confusion probabilities
are computable. We also give a mathematically more appealing form
of the universal law.
\end{abstract}

\section{Introduction}
Shepard \cite{Sh87} has put forward a
``Universal Law of Generalization''
as one of the few general psychological results governing human
cognition. The law states that the probability of
confusing two items, $a$ and $b$,
is a negative exponential function of the distance $d(a,b)$ between them in
an internal psychological Euclidean space. A drawback of this
approach may be that the Euclidean metric is one among many possible
metrics and may be appropriate in some cases but not in others. There
exists, however, a universal cognitive metric that accounts for
all possible similarities that can intuitively be perceived. It
assigns as small a distance between two objects as any cognitive
distance will do. Thus, while the positive and negative of the same picture
are far away from each other in terms of Euclidean distance, they
are at almost zero distance in terms of universal distance since
interchanging the black and white pixels transforms one picture
into the other. The universal cognitive metric, also called ``information
distance'', is a mathematical notion resulting from
mathematical logic, computer science, information theory,
and the theory of randomness. It is an ``ideal'' notion in the sense
that it ignores the limitations on processing capacity of the cognitive
system. Nonetheless, we show the following practical generalization of
the universal law [of generalization]:  if we randomly pick items
$a$ and $b$, where we allow the most complex objects,
then with overwhelming probability the universal law
of generalization holds with the internal psychological space
metric being the information metric.

\subsection{The Universal Law of Generalization}
Although intended to have broader application, the law is primarily
associated with a specific experimental paradigm---the identification
paradigm. In this paradigm, human or animal agents are repeatedly
presented with stimuli concerning a (typically small)
number of items. We denote the items themselves as $a,b, \ldots$,
the corresponding stimuli as $S_a, S_b , \ldots$, and the corresponding
responses as $R_a, R_b, \ldots$. The agents have to learn to
associate a specific, and distinct, response with each item---a response
that can be viewed as ``identifying'' the item concerned. The
stimulus $S_{a}$ is associated with item $a$ and is supposed
to evoke response $R_a$. With some probability, stimulus $S_a$ can evoke
a response $R_b$ with $b \neq a$. This means that item $b$ is
confused with item $a$.  The matrix
of $\Pr(R_a|S_{b})$ values is known as the confusability matrix. In these
terms, the universal law can be written as

\[ \Pr(R_{a}|S_{b})  \;\mbox{\rm is proportional to } e^{-d(a,b)} , \]
although we shall see below that the precise formulation is somewhat more
complex. The law is not straightforward to test, because psychological
distance $d(\cdot,\cdot)$
 can only be inferred by indirect means. Moreover, even for
the simplest sets of stimuli, such as pure tones differing in frequency,
the nature and even existence of the corresponding internal psychological
space, in terms of which distance can be defined, is highly controversial.
Shepard has, nonetheless, provided an impressive case for the universal
law.

\subsection{Shepard's Case for the Universal Law}
Shepard has argued that the technique of non-metric multidimensional scaling,
of which he is a pioneer, can be used to derive an underlying
metric  psychological
space from the confusability data itself. Specifically, the confusability
data are used to derive a rank ordering of the distances between
items on the basis of the relations between corresponding
stimuli and responses
(imposing certain assumptions, for example, to ensure that the
``distance'' between two points is symmetrical (that for all $a, b$, we have
$d(a,b) = d(b,a)$). This rank ordering is fed into a
non-metric multidimensional scaling procedure, which aims to find a way of
embedding the items into a low dimensional Euclidean space. The goal is
that the rank ordering of distances between the points should correlate as
well as possible with the rank ordering of confusabilities between items.
The underlying rationale for this procedure is that the embedding of the
items in this low-dimensional Euclidean
space can be viewed as a model of the underlying psychological space used
by the experimental participants.

Given that we have a model of the putative psychological space, and
hence can measure the distance $d(a, b)$ between items in that
space, we can assess whether the probability of confusion between $a$ and
$b$ is indeed inversely related to psychological distance, as predicted
by the Universal Law.

Shepard has amassed a large and diverse body of empirical data that, when
analysed in this way, are consistent with the universal law. The diverse
set of data that conforms to the law includes confusions between linguistic
phonemes\cite{Miller55}, sizes of circles\cite{McGuire61}, and spectral
hues, in both people\cite{Ekman} and pigeons\cite{Guttman56}. This evidence
builds an impressive case for the universal law.
Shepard has, moreover, advanced a further line of argument, that aims to
provide a theoretical justification for the universal law. Although
sympathetic to this project of rational analysis \cite{Anderson90,
ChOak2000, OakCh1998}, we will note in the
Discussion that such justifications are actually better understood as
having a different target. Rather than focussing primarily on
confusability, which is the core content of the Universal Law, they focus
on the much more difficult question of generalization from past instances
to future instances---a problem of inductive reasoning. We will therefore
postpone consideration of theoretical arguments in favour of the Universal Law.

\subsection{The Universal Law: Weighing the Evidence}
How far should we be convinced by the empirical case for the Universal Law?
There are a number of points at which the case might be challenged.

The first challenge concerns the 'universality' of the Universal Law. There
appear to be large numbers of data sets \cite{Nosofsky85, Nosofsky88a,
Nosofsky88b} from
identification paradigms for which confusability appears to be a Gaussian,
rather than a negative exponential, function of psychological distance:

\[ \Pr(R_{a}|S_{b}) \; \mbox{\rm is proportional to } e^{-d(a,b)^2}. \]

Indeed, the Gaussian generalization function is so successful empirically
that it is central to a widely-used class of exemplar models
\cite{Nosofsky85, MedShaff78}.

The empirical picture is complex, but one plausible reconciliation of the
Universal Law with apparent examples of Gaussian confusability, is that the
Gaussian confusability originates from problems of perceptually
distinguishing the stimuli, whereas the Universal Law applies when
perceptual discrimination is not the limiting factor in performance. Ennis
\cite{Ennis89} provides a useful analysis of how perceptual noise might
interact
with generalization in accordance with the Universal Law.

A further potential issue concerns the difficulties of curve-fitting.
Comparing different classes of model for fit with a set of data is a
controversial and subtle matter and fits are frequently surprisingly
inconclusive, even when very large sets of data are available
\cite{Myung2000}. Moreover, Myung and Pitt \cite{Myung1997} have recently
argued that comparisons between models are
frequently systematically biassed because one class of models is less
restrictive than the other with respect to the class of data sets that it
can model. This can lead to the counterintuitive consequence that, using
standard statistical methods, one may be likely to conclude that the data
were generated by model class $A$ rather than $B$, irrespective of whether it
was generated by model class $A$ or $B$.
Fortunately, however, the exponential
fares well from the point of view of Myung et al.s analysis---at least in
relation to the power law (which, in this context, would hold that \[
\Pr(R_{a}|S_{b}) \; \mbox{\rm is proportional to } d(a,b)^{-c}. \] for
some positive constant $c$.) which is a natural comparison. As far as we
are aware, though, recent model comparison techniques such as those
suggested by Myung and Pitt have not been applied to confusability data.

A further possible concern, which we have touched on above, is that pinning
down the structure of internal psychological spaces is a notoriously
difficult matter, and one that can be tackled from a range of theoretical
perspectives, differing from that which Shepard adopts. Indeed, the problem
of mapping magnitudes, such as sound pressure, onto a
one-dimensional internal sensory scale (perceived loudness) has occupied the
attention of psychophysicists for a century and a half without
apparent resolution. Most famously, Fechner \cite{Fechner} argued for a
logarithmic
relationship between physical intensity and internal magnitude, whereas
Stevens \cite{Stevens61} argued for a power law relationship. Not all
theorists will be confident in relying on non-metric multidimensional
scaling of confusability matrices as the solution to all these difficulties
(see Falmagne, \cite{Falmagne86} for a review of the complexities of this
area).

Yet another concern is the assumption of symmetry which
is inherent in the use of the mathematical notion of metric
to talk about the distance between two items $a$ and $b$: $d(a,b)=d(b,a)$.
In psychological
data there seem to be genuine asymmetries across many ways of measuring
confusability.
For example, complex things tend to be confused with simple things; but
simple things are less often confused with complex things. Thus, people
misremember complex shapes as simple, wobbly street plans in terms
of right angles, peculiar and unusual colors in terms of "focal" colours
(e.g. ``mauve'' becomes ``bright red'').
This problem, however, seems to be more related to rapidly blurring
of complex details in memory, which erases complexity, rather than
an essential feature of the cognitive system. Here we simply assume
symmetry of distance between two items in psychological space.

A final concern, and the one which the present paper seeks to address, is
that the view of psychological distance as Euclidean distance in an
internal multidimensional space may be too restrictive to be applicable to
many aspects of cognition. It is typically assumed that the cognitive
representation formed of a visually presented object, a sentence or a
story, will involve structured representations (e.g., \cite{Bied87,
FodBevGarr74,FodPy88,Marr82,Minsky77, SchAb77,Ullman96}. Structured
representations can
description an object not just as a set of features, or as a set of
numerical values along various dimensions, but in terms of parts and their
interrelations, and properties that attach to those parts. Thus, in
describing a bird, it is important to specify not just the presence of a
beak, eyes, claws, and feathers, but the way in which they are spatially
and functionally related to each other. Equally, it is important to be able
to specify that the beak is yellow, the claws orange and the features
white---to tie attributes to specific parts of an object. Thus,
describing a bird, a line of Shakespeare, or the plot of Hamlet as a point
in a Euclidean multidimensional space appears to require using too weak a
system of representation.

This line of argument raises the possibility that the Universal Law may be
restricted in scope to stimuli which are sufficiently simple to have a
simple multidimensional representation---perhaps those that have no
psychologically salient part-whole structure. We shall argue, however, that
the Universal Law may nonetheless be applicable quite generally, since
all these aspects are taken into account by the algorithmic information
theory approach. This leads to a more generalized form of the
Universal Law, as well as to a mathematically more appealing and
less arbitrary form.

\section{Mathematical Preliminaries}

Shepard's article raises the question
whether psychological science has hope of formulating a law
that is comparable in scope and possibly accuracy to Newton's
universal law of gravitation. The universal law of generalization for
psychological science is an tentative candidate.
In the {\em Principia} \cite{Ne}, Newton gives a few rules governing
scientific activity. The first rule is
``We are to admit no more causes of natural things than such as
are both true and sufficient to explain the appearances.
To this purpose the philosophers say that Nature does
nothing in vain, and more is in vain when less will serve;
for Nature is pleased with simplicity, and affects
not the pomp of superfluous causes.''
Here, we generalize the ``universal law of generalization''
by essentially using Newton's maxim.

We have noted that the empirical analysis of internal psychological spaces
from experimental data has proved extremely contentious. Here, we take a
complementary approach and derive the universal law
from first principles using the novel notion of information
contents of individual objects. That is, we
motivate a measure of distance between
representations of objects on a priori grounds,
drawing on recent advances in the
mathematical theory of Kolmogorov complexity \cite{LiVi97}. It turns
out that there is a very natural, and general, measure of the
``distance''
between representations, of whatever form: the information distance.
Using this very general
measure, the Universal Law of generalization still holds, subject to quite
minimal restrictions on the process by which the experimental participant
maps stimuli onto responses in the identification paradigm.

The presentation of this section has three parts. First, we provide some
general background and also describe some basic results in Kolmogorov
complexity theory. Second, we introduce and motivate the notion of
``information distance,'' which we shall use as a fundamental measure of
psychological distance. Third, we consider the nature of the
probabilitistic process by which the partipicant maps stimuli to responses,
which generates the confusion matrix in the identification paradigm---we
shall need to make only very weak assumptions about this probabilistic
process. In the next section,
we show that, given these notions, the Universal Law
holds: confusability is a negative exponential function of distance between
representations.

\subsection{Algorithmic Information Theory}

We take the viewpoint that the set of objects we are interested
is finite or infinite but countable, just like the natural numbers.
Each object can be described by using, for example English.
That means we can describe every object by a finite string in some
fixed finite alphabet. By encoding the different letters of that alphabet in
bits (0's and 1's) we reduce every description or representation
of the object to a finite binary string. A similar argument presumably holds
for the physical manner an object is represented in an agents
cognitive system. This way we reduce the representation
of all objects that are relevant in this discussion
to finite binary strings. In the unlikely case
that there are relevant objects that cannot be so represented
we simply agree that they are not subject of this discussion.

In the psychology and cognitive literature there have been a great
number of proposals for encodings of patterned sequences and
defining the complexities of the resulting encodings. See for
example the survey in Simon \cite{Si72}. All such encodings are
special types of computable codes, which means that all of them can be decoded
by appropriate machines or programs. Mathematically  one says that
every such code can be decoded by an appropriate {\em Turing machine}:
a convenient model introduced in \cite{Tu36}
to formally capture the intuitive notion of ``computation''
in its greatest generality. It has turned out
that all different mathematical proposals to formulate a more
general notion of computability after all turned out to
be equivalent to the Turing machine. Since then, the so-called
{\em Church-Turing thesis} states that the Turing machine captures the
most universal and general notion of effective computability,
and is the formal equivalent of our intuitive notion
of the same. It is universally used in formal arguments.
There is no need to go into details here, they can be found in any
textbook on computable functions and effective processes, for example,
\cite{Od89} or a section 1.7 in \cite{LiVi97}.
What is important here is that there is a general
code that subsumes all computable codes mentioned above.
This is the code decodable
by a so-called ``universal'' Turing machine.
In effect, such a machine works with a code book that enumerates all
computable codes. By prefixing an encoded item with the index
in the enumeration of the particular code that has been used,
the universal Turing machine can decode. Clearly, this universal
encoding need  not be longer than the shortest two-part code
consisting of the index of a particular code used plus the length of the
resulting encoding. Stating that the universal code can be decoded
by a universal Turing machine is equivalent to that it
is a program in a universal programming language like C++ or Java.
This leads to a notion of information content of
an individual object pioneered by the Russian mathematician
A.N. Kolmogorov \cite{Ko65}. This notion should
be contrasted  to Shannon's statistical
notion of information \cite{Sh48} which deals with the average number
of bits required to communicate a message from a probabilistic ensemble between
a sender and a receiver.
In this paper we keep the discussion informal; an introduction, epistimology
and rigorous
treatment of the theory is given in \cite{LiVi97}.

The {\em Kolmogorov complexity} $K(x)$ of a  finite object
$x$, is defined as the
length of the shortest binary computer program that produces $x$ as an
output.\footnote{Strictly, it is important that the program is a prefix
program---that
is, that no initial segment of the binary string comprising the program
itself defines a valid program; and, equally, that no non-trivial
continuation of the binary string comprising the program defines a valid
program. The restriction to prefixes ensures that, for example, given a
binary string that corresponds to a concatenation of programs, there is no
ambiguity concerning how the string should be divided into discrete
programs. Although tangential to the discussion here, the use of prefix
complexity is of considerable technical importance \cite{LiVi97,ViLi00}.}
 Thus, objects
such as a string of one billion `1's or, a binary code for a digitized
picture of an untextured rectangle, or the first million digits of
$\pi = 3.1415 \ldots$ are
reasonably simple, because there are short programs that can generate these
objects. On the other hand, a typical binary sequence generated by tossing
a coin is complex---the sequence is its own shortest program, because there
is no hidden structure that can be used to find a shorter code.
Kolmogorov complexity theory, also known as algorithmic information theory,
is a modern notion of randomness dealing with the quantity of information
in individual objects.
The Kolmogorov complexity of an object is a form of
absolute information of the individual object, in contrast to
standard (probabilistic) information theory  \cite{CT91} which is only
concerned with the average information of a random source.

The definition of Kolmogorov complexity may appear to be rather specific.
But this appearance is misleading. For example, The restriction to a binary
coding alphabet can easily be dispensed with---switching to an alphabet with
$n$ letters amounts merely to rescaling all Kolmogorov complexities by a
multiplicative constant,\footnote{Specifically,
this constant is $\log n$. All logarithms in this
paper are binary logarithms unless otherwise
noted.} but has no other impact.
The binary alphabet is used by convention, to provide a fixed measuring
standard. More interestingly, it might appear that the length of the
shortest program that generates a specific code must inevitably be relative
to the choice of programming language. But a central result of Kolmogorov
complexity theory, the Invariance Theorem \cite{LiVi97}, states that
the shortest description of any object is invariant (up to a constant)
between different universal languages.  Therefore, it does not matter
whether the universal language chosen is C++, Java or Prolog---the length of
the shortest description for each object will be approximately the same.
Let us introduce the notation $K_{C++}(x)$ to denote the length of the shortest
C++ program which generates object $x$; and $K_{Java}(x)$
to denote the length of
the shortest Java program.  The Invariance Theorem implies that $K_{C++} (x)$
and $K_{Java}(x)$ will only differ by some constant, $c$, (which may be
positive
or negative) for {\it all} objects $x$, including, of course, all possible
perceptual stimuli.  Formally, there exists a constant $c$ such that for
all objects $x$:

$$|K_{C++}(x) - K_{Java}(x)| \leq  c.$$
Thus, in specifying the complexity of an object, it is
therefore possible to abstract away from the particular language under
consideration.  Thus the complexity of an object, $x$, can be denoted
simply as $K(x)$---referring to {\it the} Kolmogorov complexity of that object.

Why is Kolmogorov complexity language invariant? To see this intuitively,
note that any universal language can be used to encode any other universal
programming language. This follows from the preceding discussion because a
programming language is just a particular kind of computable mapping, and
any {\it universal}
programming language can encode any computable mapping. For example,
consider two universal computer languages which we call ``C++'' and ``Java.''
Starting with C++, we can write a program, known in computer science as a
compiler, which translates any program written in Java into C++.  Suppose
that this program has length has length $c_1$. Suppose that we know
$K_{Java}(x)$, the length of the shortest program which generates an
object $x$ in
Java. What is $K_{C++}(x)$, the shortest program in C++ which encodes $x$?
Notice
that one way of encoding $x$ in C++ works as follows---the first part of the
program  translates from Java into C++ (of length $c_1$), and the second part
of the program, which is an input to the first, is simply the shortest Java
program generating the object. The length of this program is the sum of the
lengths of its two components: $K_{Java}(x) + c_1$. This is a C++ program which
generates $x$, if by a rather roundabout means.  Therefore $K_{C++}(x)$, the
shortest possible C++ program must be no longer than this: $K_{C++}(x)
\leq  K_{Java}(x) + c_1$.
An exactly symmetric argument based on translating in the
opposite direction establishes that: $K_{Java}(x) \leq   K_{C++}(x) + c_2$.
Putting these
results together, we see that $K_{Java}(x)$ and $K_{C++}(x)$ are the same
up to a
constant, for all possible objects $x$. This is the Invariance Theorem
that
Kolmogorov complexity is language invariant.

The implication of the Invariance Theorem is that the functions $K( \cdot)$
(and $K( \cdot| \cdot)$, that we introduce below)
though defined in terms of a
particular programming language, are language-independent up to an additive
constant
 and acquire an asymptotically universal and absolute character
through  the Church-Turing thesis, from the ability of universal machines to
simulate one another and execute any effective process.
The Kolmogorov complexity of a string can be viewed as an absolute
and objective quantification of the amount of information in it,
giving a rigorous formal and most general notion corresponding
to our intuitive notion of shortest effective description length.
This may be called {\em Kolmogorov's thesis}.
   This leads to a theory of {\em absolute} information {\em contents}
of {\em individual} objects in contrast to classical information theory
which deals with {\em average} information {\em to communicate}
objects produced by a {\em random source}.
  Since the former theory is much more precise, it is perhaps surprising that
analogs of many central theorems in classical information theory
nonetheless hold for Kolmogorov complexity, although in a somewhat weaker
form.

We have mentioned that shortest code length is invariant for {\it
universal} programming languages. How restrictive is this? The constraint
that a system of computation is universal turns out to be surprisingly
weak---all manner of computation systems,
from a simple automaton with under 100
states supplied with unlimited binary tape from which it can read and write,
to numerous word processing packages, spreadsheet and statistical
packages, turn out to define universal programming languages. It seems that
universality is likely to be obeyed by a computational system as elaborate
as that used involved in cognition.

The basic notion of Kolmogorov complexity has been elaborated into a rich
mathematical theory, with a wide range of applications in mathematics and
computer science. It has also been applied in a range of contexts in
psychology, from perceptual organization (see \cite{Chater96, Helm} for
different uses of the theory), to psychological judgements of
randomness \cite{FalkKonald}, to providing the basis for a theory of
similarity \cite{ChaHahn97}. Indeed, the idea that cognition seeks
the simplest explanation for the available data, inspired by results in
Kolmogorov complexity, has even been suggested as a fundamental principle
of human cognition \cite{Chater97, Chater99}.

\subsection{Information distance}

Kolmogorov complexity is defined for a single object, $x$. But an immediate
generalization, conditional Kolmogorov complexity, $K(y|x)$ provides a
measure of the degree to which an object $y$ differs from another object $x$.
$K(y|x)$ is defined as the length of the shortest program (in a universal
programming language, as before) that takes $x$ as input, and produces $y$ as
output. The intuitive idea is that if items are distant from each other,
then it should require a complex program to turn one into the other.
At this point it is useful to recall the mathematical
formulations of the notions of ``distance'' and ``metric'':

A {\em distance} function $D$ with nonnegative
real values, defined on the Cartesian product $X \times X$ of
a set $X$ is called a {\em metric}
on $X$ if for every $x,y,z \in X$:
\begin{itemize}
\item
$D(x,y)=0$ iff $x=y$ (the identity axiom);
\item
$D(x,y)+D(y,z) \geq D(x,z)$ (the triangle inequality);
\item
$D(x,y)=D(y,x)$ (the symmetry axiom).
\end{itemize}
A set $X$ provided with a metric is called a {\em metric space}.
For example, every set $X$ has the trivial {\em discrete metric}
$D(x,y)=0$ if $x=y$ and $D(x,y)=1$ otherwise. All information
distances in this paper are defined on the set $X=\{0,1\}^*$ (that is, the
set of all finite strings composed of 0s and 1s)
and satisfy
the metric conditions up to an additive
constant or logarithmic term while the identity axiom can be
obtained by normalizing.

The conditional complexity function $K(y|x)$ trivially obeys
identity, because no program at all is required to transform an item into
itself.\footnote{Note that, throughout, due to language invariance,
Kolmogorov complexities are only specified up to an additive constant. So,
in a particular language, $K(x|x)$ could be non-zero---if, for example, some
instructions are required to implement the 'null' operation (this is
typically true of real programming languages, in which the null string is
not treated as a valid program. But the length of this program will, by
language invariance, be bounded by a constant, for all possible $x$.}
Conditional complexity also obeys the triangle inequality: $K(x|z) \leq
K(x|y) + K(y|z)$. This follows immediately from the observation that the
concatenation of a program mapping $z$ into $y$ (with minimum length $K(y|z)$),
and a program mapping $y$ into $x$ {\it is} (with minimum length $K(x|y)$).
Using
this concatenation, it is clearly possible to map $x$ to $z$ using a program of
length no more than the sum of these individual programs: $K(x|y) + K(y|z)$.
This sum must therefore be at least as great as the length shortest program
mapping from $z$ to $x$, that is $K(x|z)$,
where $K(x|z)$ is typically smaller by there being shorter
programs which perform this mapping without going through the intermediate
stage of generating $y$. Thus, the triangle inequality holds for
$K(\cdot|\cdot)$.

But, as it stands, $K(y|x)$ is not appropriate as a distance measure, because
it is asymmetric. Consider the null string $\epsilon$.
$K( \epsilon|x)$ is small, for every $x$,
because to map the input $x$ onto the null string simply involves deleting
$x$, which is a simple operation. Conversely, $K(x|\epsilon) = K(x)$,
which
can have any value whatever, depending on the complexity of $x$. Symmetry can
be restored by, for example, taking the sum of the complexities in both
directions: $K(x|y) + K(y|x)$, or alternatively, the maximum of both
complexities $\max \{K(x|y),K(y|x)\}$. It is easy to verify that the resulting
measures, known as {\em sum distance} and {\em max distance}, respectively,
qualify as distance metrics \cite{BGLVZ98,LiVi97}. For example, the sum
distance and the max-distance between
$x$ and the null string $\epsilon$ are given by
$K(x|\epsilon ) + K(\epsilon|x) = K(x)
= \max \{K(x|\epsilon ) , K(\epsilon|x)\}$.

Max and sum-distances are close but not necessarily equal.
Denoting sum and max distance
respectively by $D_{\mbox{\footnotesize \rm sum}}$ and $D_{\max}$,
it is easy to verify that, for every $x, y$:

\begin{equation}\label{*}
 D_{\max}(x, y) \leq D_{\mbox{\footnotesize \rm sum}}(x, y)) \leq 2
D_{\max}(x, y) .
\end{equation}

For the present purpose of putting the Universal Law on a
formal mathematical footing, it is important to consider the epistimological
motivation of these distances.  The {\em information distance}
is defined in \cite{BGLVZ98}
as the length of a shortest binary program that
computes $x$ from $y$ as well as computing $y$ from $x$.
  Being shortest, such a program should take advantage of any
redundancy between the information required to go from $x$ to $y$
and the information required to go from $y$ to $x$.
  The program functions in a catalytic capacity in the
sense that it is required to transform the
input into the output, but itself remains
present and unchanged throughout the computation.
Note that while a program of length $K(x|y)+K(y|x)$ by
definition can compute from $y$ to $x$ (a subprogram
of length $K(x|y)$) and from $x$ to $y$ (a subprogram
of length $K(y|x)$), it is by no means clear (and happens to be false)
that such a program is necessarily the shortest that performs both the mapping
from $x$ to $y$ and the mapping from $y$ to $x$.
  A ($K(x|y)+K(y|x)$)-length program is not minimal if the information
required to compute $y$ from $x$ can be made to overlap with that
required to compute $x$ from $y$.

  In some simple cases, {\em complete} overlap can be achieved, so that
the same minimal program suffices to compute $x$ from $y$ as to
compute $y$ from $x$.
We first need an additional notion. A binary string $x$ of $n$ bits
is called {\em incompressible} if $K(x) \geq n$. A simple argument
suffices to show that the overwhelming
majority of strings is incompressible, \cite{LiVi97}.
We continue the main argument.
  For example if $x$ and $y$ are independent incompressible binary strings of
the same length $n$ (up to additive
constants we have $K(x|y),K(y|x)\geq n$), then
their bitwise exclusive-or $x \oplus y$ serves as a minimal program
for both computations.  (If $x=01011$ and $y=10001$, then
$z=x \oplus y = 11010$. Since $z\oplus y =x$ and $z \oplus x = y$
we can use $z$ as a program both to compute from $y$ to $x$ and
to compute from $x$ to $y$.)

  Similarly, if $x=uv$ and $y=vw$ where $u$, $v$, and $w$ are
independent incompressible strings of the same length, then $u \oplus w$
along with a way to distinguish $x$ from $y$ is a
minimal program to compute either string from the other.
  Now suppose that more information is required for one of these
computations than for the other, say,
 \[
  K(y|x) > K(x|y).
 \]
  Then the minimal programs cannot be made identical because they
must be of different sizes.
  In some cases it is easy to see that the overlap
can still be made complete, in
the sense that the larger program (for $y$ given $x$) can be made to
contain all the information in the shorter program, as well as some
additional information.
  This is so when $x$ and $y$ are independent incompressible strings of
unequal length, for example $u$ and $vw$ above.
  Then $u\oplus v$ serves as a minimal program for $u$ from $vw$, and
$(u \oplus v)w$ serves as one for $vw$ from $u$.

 A principal result of \cite{BGLVZ98} shows that,
up to an additive  logarithmic error term,
the information required to translate
between two strings can be represented in this maximally overlapping
way in {\em every case}. That is, the minimal program to translate back and
forth
between $x,y$ has length not larger than $\max \{K(x|y),K(y|x)\}$.
It is straightforward
that the minimum length program to do this back and forth translation
cannot be shorter, since by definition of
Kolmogorov complexity the translation in direction $x$ to $y$ requires
a program of length at least $K(y|x)$ and the translation
in the direction of $y$ to $x$ requires a program of length at
least $K(x|y)$. Therefore, the length of the shortest binary program that
translates back and forth between two items is
called the {\em information distance} between the two items,
and it is equal to $D_{\max} (x,y)$---to be
precise, up to an additive logarithmic term which we ignore in  this
discussion.

Max-distance has a particularly attractive universal quality:
it is, in a sense, the {\it minimal} distance measure, in a broad class of
distance measures that might be termed ``computable,'' as we now see.

We say that a function from a discrete domain to the reals
(for example a distance metric)
is {\em semicomputable from
above} if it can be
approximated from above by some computable process. This is a very
weak condition. For example, it is weaker than the assumption that a
function is computable. It requires merely that there is some
computable process that outputs a sequence of successive approximations to
the function value, that are successively decreasing, and which
converge to be as close as desired to the distance metric, given sufficient
computation\footnote{It does not require, for example, that it is possible
to actually output the ``correct'' distance values---or, indeed, to announce
the degree of approximation that has been achieved after a given amount of
computation.}. If we assume the Church-Turing thesis, that human cognition
can encompass only computable processes, then it seems that this assumption
follows automatically.

To make sense of the notion of a ``minimal'' distance measure, we need some
normalization condition, to fix the ``scale'' of the distances. Without such
a condition, we could simply divide the values given
by a distance metric by an
arbitrarily large constant $c$ to get a more ``minimal'' distance metric.

 For a cognitive similarity metric the metric requirements
  do not suffice: a distance measure like $D(x,y) = 1$
for all $x \neq y$ must be excluded.
  For each $x$ and $d$, we want only finitely many elements $y$ at a
distance $d$ from $x$.
  Exactly how fast we want the distances of the strings $y$ from $x$
to go to $\infty$ is not important: it is only a matter of scaling.
  In analogy with Hamming distance in the space of binary sequences,
it seems natural to require that there should not be more than $2^d$
strings $y$ at a distance $d$ from $x$.
  This would be a different requirement for each $d$.
  With prefix complexity, it turns out to be more convenient to
replace this double series of requirements (a different one for each
$x$ and $d$) with a single requirement for each $x$:
 \[
  \sum_{y: y \neq x} 2^{-D(x,y)}<1 .
 \]
  We call this the {\em normalization property} since a certain sum is
required to be bounded by 1.

  We consider only distances that are computable in some broad sense.
  This condition will not be seen as unduly restrictive.
  As a matter of fact, only upper-semicomputability of $D(x,y)$ will
be required.
  This is reasonable: as we have more and more time to process $x$ and
$y$ we may discover more and more similarities among them, and thus
may revise our upper bound on their distance.
  The upper-semicomputability means exactly that $D(x,y)$ is the limit
of a computable sequence of such upper bounds.

\begin{definition}
  An {\em admissible distance} $D(x,y)$ is a total nonnegative
function on the pairs $x,y$ of binary strings that is 0 if and only if
$x=y$, is symmetric, satisfies the triangle inequality, is
upper-semicomputable and normalized, that is, it is an upper-semicomputable,
normalized, metric.  An admissible distance $D(x,y)$
is {\em universal} if for every admissible distance $D'(x,y)$ we have
$D(x,y) \leq D'(x,y) +c_D$ where $c_D$ may depend on $D$ but not on $x$ or $y$.
\end{definition}

  In \cite{BGLVZ98} a remarkable theorem shows that $D_{\max}$ is a universal
(that is, optimal) admissible distance. Formally, every admissible
distance metric $D$  has an associated constant $c$ such that

\[ D_{\max} (x, y) \leq D(x, y)+ c , \]
for every $x$ and $y$.

  As already discussed above,
the universal distance $D_{\max}$ happens to also have a
``physical'' interpretation as the approximate length of the
the smallest binary program that transforms $x$ into
$y$ and vice versa.
That is, for all the infinitely many $x, y$, and hence
the infinite number of distances between them, the $D_{\max}$ distance
is never more than a
finite additive constant term greater than the corresponding $D$-distance
with respect to any admissible distance metric $D$, where the additive
constant may depend on $D$ but is independent of $x$ and $y$.
\footnote{By (\ref{*}) $D_{\mbox{\footnotesize \rm sum}} (x,y) \leq 2
D(x,y)+2c$,
because the two measures $D_{\max}$ and $D_{\mbox{\footnotesize \rm sum}}$
are within a constant factor 2 of each other.}

Intuitively, the significance of this is that
  the universal admissible distance minorizes {\em all}
admissible distances: if two pictures are $d$-close under some
admissible distance, then they are $d$-close up to a fixed additive
constant under this universal
admissible distance.
  That is, the latter discovers all effective feature similarities
or cognitive similarities
between two objects: it is the universal cognitive similarity metric.
The remarkable thing about information distance measures such as $D_{\max}$
 is that, with respect to the class of computable distance measures
(subject to the normalization condition described above), they are minimal.
That is, if any computable measure treats two items as near, then
information distance measures will also treat the items as `reasonably'
near.

The typical distance
measures considered in psychology, artificial intelligence or
mathematics are {\it not} universal. This is because they favor some
regularities among the items that consider, but entirely ignore other
regularities---and some of these regularities may be the basis of computable
(and hence allowable) distance measures.
  Let us look at some examples.
 Identify digitized black-and-white pictures with binary strings.
  There are many distances defined for binary strings.
  For example, the Hamming distance and the Euclidean distance.
The Hamming distance between two $n$-bit vectors is the number
of positions containing different bits; the Euclidean distance
between two $n$-bit vectors is the square root of the Hamming distance.
  Such distances are sometimes appropriate.
  For instance, if we take a binary picture, and change a few bits on
that picture, then the changed and unchanged pictures have small
Hamming or Euclidean distance, and they do look similar.
  However, this is not always the case.
  The positive and negative prints of a photo have the largest
possible Hamming and binary Euclidean distance, yet they look similar to us.
  Also, if we shift a picture one bit to the right, again the Hamming
distance may increase by a lot, but the two pictures remain similar.
As another example, a metric of
similarity based on comparing overlap of features \cite{Tversky77} will
treat items that have precisely opposite patterns of features as very
distant. But, of course, with respect to the $D_{\max}$ measure
such items are very close since the program saying "take the opposite
of every feature" suffices to change one item into the other.
Hence, such a feature-based metric
is not a minimal distance. Similarly, if items are represented as real-valued
vectors, and the Euclidean distance metric is used, then items corresponding
to vectors ${\bf v}$ and $2{\bf v}$  will have distance equal to
the Euclidean length of ${\bf v}$, while $D_{\max}$ is small.

We believe that, in the present context, the minimality of information
distance is a substantial virtue, because minimal distance measures make
the least commitment to the specific similarity metric used by the
cognitive system---because they approximate all possible (computable)
metrics. Thus, minimal distance measures seem the ideal candidate ``null
hypotheses'' about the structure of psychological similarity.

We have considered some technical reasons why measures based on information
distance are attractive general distance measures. These measures gain some
additional psychological interest because of its relation to the recently
proposed Representational Distortion theory of psychological similarity
\cite{ChaHahn97}. According to Representation Distortion, the
psychological similarity of two items depends on the complexity of the
transformation required to ``distort'' the representation of one of the items
into a representation of the other item. The notion of complexity is then
assumed to be related to the notion of conditional Kolmogorov complexity,
as described here. According to this viewpoint, the flexibility of measures
like information distance is appropriate because it reflects the
flexibility of the cognitive system---to choose arbitrary ways of
interrelating, aligning and connecting representations, rather than being
constrained to use a fixed similarity measure. This account of similarity,
although early in its development, has received some empirical support
\cite{HCR2000}.

\subsection{How Items are Confused}

We assume a very general, and weak, model of similarity---based on
information distance. We next need a general model of the how items are
confused with each other. Fortunately, only a very weak assumption is
required. First, we assume that there is a discrete set of items
$a$, stimuli $S_a$, and responses $R_a$. Moreover, these are
associated with one another in the sense that there is a fixed program
that on input $x$ computes $y$ where $x$ and $y$ are choosen from among
of $a,S_a,R_a$.
Secondly, for each stimulus $a$, the probability distribution
$\Pr(R_b|S_a)$ over the different responses, $b$, is itself semicomputable
from below. That is,
it can be approximated
from below  by a computable process that produces a
monotonically increasing series of approximations to $\Pr(R_b|S_a)$ which
approach arbitrarily closely, given sufficient computing time. This is a
weaker condition, of course, than the condition that the probability
distribution can be actually be computed exactly by some computable
process---which is equivalent to the distribution being both
semicomputable from above and from below.
Recall that the celebrated Church-Turing Thesis, see for example
\cite{Od89}, states that everything
which is intuitively computable can be computed formally by a
Turing machine, or, equivalently, a standard computer supplied with
a large anough memory.
Assuming the
Church-Turing thesis  implies that processes executed
by the cognitive system are
computable functions. In particular, therefore, this condition will include
any computational account of the process by which stimuli $S_a$ are mapped
onto responses $R_b$.

\section{The General Universal Law of Generalization}
Having outlined the notion of information distance, and provided a weak
condition on the cognitive processes by which confusability between items
occurs, we are now in a position to show how the generalized
``algorithmic'' version of the Universal Law can be
derived from first principles.

The idea behind the proof is to place bounds on the confusability
probabilities, $\Pr(R_b|S_a)$,
simply in virtue of its semicomputability, using basic
results of Kolmogorov complexity theory. These bounds can be interpreted as
providing a direct connection between confusability and the measure of
information distance.
We use standard results from Kolmogorov complexity theory, which
provide the bounds on $\Pr(R_b|S_a)$ that we require.

\subsection{Optimal Codes and Entropy}
For technical reasons  we recall some notions
from information theory \cite{CT91}.
Suppose we have a random source emitting letters from the alphabet
with certain frequencies. Our task is to encode messages consisting
of many letters in binary in such a way that on average
the encoded message is as short as possible.
It is evident that by assigning the few shortest binary sequences
to the most common letters and the longer sequences to the rare
ones, the expected length of a message
is less than if we assigned equal length codes to all letters.
Thus, the Morse code in telegraphy is adapted to the
frequency of letter occurrences in English. It assigns short sequences
of dots and dashes to more frequently occurring letters:
``a'' is encoded as ". -" and  ``t'' is encoded as ``-''.
Long sequences of dots and dashes are assigned
to less frequently occurring letters such as ``z'' which is encoded as
``- - . .''.
A {\em prefix code} has the property that no code word starts
with another code word as proper initial segment (prefix). This
property makes it possible to parse an encoded message into the
sequence of code words from which it is composed in only one way:
We can unambiguously retrieve the encoded message.
Note that the Morse code is not a prefix-code. A prefix code for
the letters a,b, $\ldots$ ,z is, for example, to encode ``a''
by ``.-'', the letter ``b'' by ``..-'', and so on. This example is
not very efficient; it is essentially a tally code. It is easy
to design more efficient prefix-codes. Nonetheless, since prefixes
are excluded, it is clear that prefix-codes cannot be as concise
as general codes. But prefix-codes have a very general and central
property that makes them more practical than other codes: for every
code that is uniquely decodable there is a prefix-code that has
precisely the same lengths of code words. Thus, when we want
unambiguous codes then we can as well restrict ourselves to prefix-codes:
they are uniquely decodable and have the additional advantage
that we can parse them in one pass going left-to-right.
Moreover, it is well-known that there is a tight connection between
prefix codes, probabilities, and notions of optimal codes:
Call the letters to be encoded by the name ``source words''.
Consider an ensemble of source letters
with source word $x$ having probability $P(x)$.
Assign code words with code word length $l_P(x)$
to source word $x$.
The so-called Noiseless Coding Theorem of C. Shannon
states that
among all prefix codes
the minimal average code word length,
the average taken with respect to the distribution $P$, satisfies
\[ H(P) \leq \sum_x P(x)l_P (x) \leq H(P)+1 \]
where $H(P) = - \sum_x P(x) \log P(x)$ is called the {\em entropy} of $P$.
This minimum is reached by the so-called
Shannon-Fano  code (the details of which do not matter here)
where we assign a code word of length $- \lceil \log P(x) \rceil$
to source word $x$.
Intuitively, this code is
optimally ``adapted'' to the probability distribution $P$ of the source
words.

\subsection{The Kolmogorov Code}
A trivial application of this result, generalized
to conditional probability, is that using a code that is well-adapted to
probability distribution $\Pr ( \cdot | S_a)$, the Shannon-Fano code length
of $R_b$, given $S_a$, is
$-\log \Pr(R_b|S_a)$.
This allows us to quantify the code length of a particular way of mapping
$S_a$ onto $R_b$. First, specify the probability
distribution $\Pr$---this can be done
using a computer program of length $K(\Pr)$ (the existence of such a computer
program is guaranteed by the condition that $\Pr$ is computable). Then
specify $R_b$, given $S_a$ using the probability distribution $\Pr$, which
takes length $-\log \Pr(R_b|S_a)$ using the Shannon-Fano code.
Thus, the total length of this way of
mapping from $S_a$ to $R_b$ is: $K(\Pr)  -\log \Pr(R_b|S_a)$. Obviously,
every
computable code that maps $S_a$ to $R_b$ must be at least as
long as the shortest computable code which does this, which length is, by
definition, $K(R_b|S_a)$. Thus, we can infer that:

\[ K(R_b|S_a) \leq K(\Pr)  -\log \Pr(R_b|S_a), \]

which, when rearranged,
provides an {\em upper bound} on $\Pr(R_b|S_a)$:

\[ \Pr(R_b|S_a) \leq 2^{K(\Pr)-K(R_b|S_a)} . \]

In the following it is convenient to use a special
notation for (in)equality up to an additive constant.
  From now on, we will denote by $\lea$ an inequality to within an
additive constant, and by $\eqa$ the situation when both $\lea$ and
$\gea$ hold.

We derive a {\em lower bound}: Suppose we sample from a
distribution $\Pr$, and encode the outcomes using an optimally adapted code,
as described above. We can then write down the expected code length as

\begin{eqnarray*}
E_{\Pr} (- \log \Pr(x)) & \eqa &
\sum_x \Pr(x) (- \log \Pr(x)) \\
& \eqa & - \sum_x \Pr(x) \log \Pr(x).
\end{eqnarray*}

Here $E_{\Pr} f(x) = - \sum_x \Pr(x) f(x)$
 is called the {\em expectation} of $f(x)$ with respect to $\Pr$.
With $f(x) = - \log \Pr(x)$ this is the above
expression for the {\it entropy} of $\Pr$. Now suppose that we consider,
instead, the expected value of the Kolmogorov complexity of $x$---the
shortest code length for $x$, in a universal programming language. In
general, of course, this will be at least as great as the entropy---because
the entropy reflects the shortest
expected code length for $x$, using a code which is
optimally adapted to $\Pr$. So this means that

\[ E_{\Pr} (-\log \Pr(x)) \leq E_{\Pr} K(x) . \]

Nonetheless, though, there will typically be individual values of $x$ for
which Kolmogorov complexity is significantly
 less than the code length ($\approx - \log \Pr(x)$) optimized to $\Pr$.
For example, suppose that $\Pr$ is an extremely simple
distribution over binary strings, such that 0 and 1 values both have a
probabilty of .5, and are independent---as if, for example, the string were
generated by a series of fair coin flips. Consider the string that consists
of a million consecutive 1s. According to $\Pr$, the probability of this
string is $2^{-1,000,000}$, and the code length according to the code
optimally adapted to $\Pr$ is $-\log 2^{-1,000,000} = 1,000,000$. Indeed, this
same code length will be assigned for every binary string of $1,000,000$
characters generated by $\Pr$, because according to $\Pr$ all such strings have
the same probability of occurring. However, the Kolmogorov complexity of
this particular string will, of course, be considerably less than $1,000,000$
bits---because a short computer program can print a million 1s and then halt.

The reason that this particular string generated by $\Pr$ has a smaller
Kolmogorov complexity that is associated with the optimal code for $\Pr$, is
that the string has some additional structure, that is unexplained by $\Pr$.
The existence of this additional structure (such as being a sequence of
repeated items, or alternating items, or encoding $\pi = 3.14 \ldots$ in
binary, or whatever
it may be) can therefore be used to provide an unexpectedly short code for
the string. Intuitively, though, it seems that strings generated by $\Pr$ with
such additional useful structure must be rare---it would seem likely that the
overwhelming majority of strings generated by $\Pr$ will merely be typical of
the distribution, and hence will not contain any useful ``unexpected''
structure. The Kolmogorov complexity of these items will, therefore, be at
least as great as the code length according to the code optimally adapted
to $\Pr$. This intuition is indeed correct. It can be shown that the
probability that an item, $x$, drawn from $\Pr$, is such that

\[ -\log \Pr(x) \leq K(x) \]

goes to 1 for length of $x$ grows unboundedly \cite{LiVi97,ViLi00}.
That is, almost all probability is concentrated on items $x$ satisfying this
inequality---and if the probability is not dramatically skewed this implies
that the overwhelming majority of $x$'s do so.
Items for which this
inequality holds are known as $\Pr( \cdot )$-random,
indicating that they do
not have sufficient 'unexpected' structure to support an shorter coding
than would be expected from $\Pr$\footnote{Here, we touch on the more general
idea that the randomness of a string may be assessed by considering its
Kolmogorov complexity. This idea has been developed into a deep
mathematical theory of 'algorithmic' randomness.
The common meaning of a ``random object'' is an outcome
of a random source. Such outcomes have expected properties
but particular outcomes may or may not possess these expected
properties. In contrast, we use
the notion of randomness of individual objects. This elusive notion's long
history goes back to the initial attempts by von Mises,
\cite{Mi19}, to formulate
the principles of application of the calculus of probabilities to
real-world phenomena.
Classical probability theory
cannot even express the notion of ``randomness of individual objects.''
Following almost half a century of unsuccessful attempts,
the theory of Kolmogorov complexity, \cite{Ko65}, and Martin-L\"of tests
for randomness, \cite{ML66}, finally succeeded in formally
expressing the novel notion of
individual randomness in a correct manner, see \cite{LiVi97}.
 Every individually random object
possesses individually all effectively
testable properties that are only expected for outcomes of
the random source concerned. It will
satisfy {\em all} effective tests for randomness---
known and unknown alike. Details are beyond the scope of this treatment,
but see the discussions in  \cite{ML66,LiVi97}.
}.

A straightforward generalization of this result to conditional probability,
and its application in the present context yields the result that, for the
$R_b$ that are $\Pr(.|S_a)$-random (and the probability of sampling such an
item
from $\Pr( \cdot |S_a)$ will be almost 1), then

\[ - \log \Pr(R_b|S_a) \leq K(R_b|S_a) . \]

This equation can be rearranged to give a lower bound on $\Pr(R_b|S_a)$:

 \[ 2^{-K(R_b|S_a)} \leq \Pr(R_b|S_a) \]

Putting the upper and lower bounds together, we can conclude that, for
$\Pr( \cdot|S_a)$-random items:

\[
2^{-K(R_b|S_a)} \leq \Pr(R_b|S_a) \leq 2^{K(Pr)-K(R_b|S_a)} .
\]

This result implies that, for allmost all items (the
$\Pr( \cdot |S_a)$-random
items), $\Pr(R_b|S_a)$ is close to $2^{-K(R_b|S_a)}$, to within a
multiplicative
factor, $2^{K(\Pr)}$. Since $K(\Pr)$ is constant, independent of the
items $a$ and $b$ we can simplify the formulas,
using the earlier introduced notation ``$\eqa$'', to
\begin{equation}\label{**}
\log \Pr(R_b|S_a) \eqa -K(R_b|S_a),
\end{equation}
for allmost all items $b$ (the
$\Pr( \cdot |S_a)$-random
items) with respect to every item $a$. That is, for almost all
pairs of items $a,b$ with $\Pr (\cdot|a)$-probability
going to 1 for $b$ increasing
with every fixed $a$.

\subsection{Formal Derivation of the Law}
Now we are in a position to directly relate Shepard's Universal Law to
information distance. Shepard uses a specific measure, $\G(a, b)$, as a
measure of what he terms the `generalization'
between items $a$ and $b$. Here $S_a$ is the stimulus related to
item $a$ with the
correct corresponding response $R_a$. Possibly, the stimulus
$S_a$ elicits another response $R_b$ ($b \neq a$). The probability
of this happening is $\Pr (R_b | S_a)$.

\begin{equation}\label{++}
\G(a,b)= \left[ \frac{ \Pr(R_a|S_b)Pr(R_b|S_a)}{ \Pr(R_a|S_a)Pr(R_b|S_b)}
\right]^{\frac{1}{2}}
\end{equation}

To express $\G(a,b)$ in terms of Kolmogorov complexity observe
the following.
We have assumed at the outset that there is  a simple fixed program,
of length say $C$ bits,
that maps $S_x$ to $R_x$ for all $x$'s.
This means  that $K(R_a|S_a)$ and $K(R_b | S_b)$
are upper bounded by a fixed constant $C$ independent of
variable items $a$ and $b$. Moreover,  $K(R_a|S_a)$ and $K(R_b | S_b)$
are strictly positive as a consequence of the definition of Kolmogorov
complexity (the Universal Turing Machine must have some program to
do the transformation). Therefore, the denominator in (\ref{++})
can be replaced by a positive constant independent of $a$ and $b$.
Taking this into account, and
substituting (\ref{**}) into (\ref{++}) we obtain that, for almost all $a, b$
(the almost all
$\Pr( \cdot |S_a)$-random
items $b$ with respect to every item $a$
%
%
in the above sense of concentration
of $\Pr$-probability),
\begin{equation}\label{@}
\log \G(a,b) \eqa \frac{1}{2} \left[-K(R_a|S_b) - K(R_b|S_a) \right] .
\end{equation}

We have also assumed at the outset that there are fixed length
programs that computes $S_a$ from $a$, $R_a$ from $a$,
$S_a$ from $R_a$, and so on, for every item $a$.
Therefore, $K(R_b|S_a) \eqa K(b|a)$
and $K(R_a|S_b) \eqa K(a|b)$.
Earlier, we defined the ``sum''-information distance
$D_{\mbox{\footnotesize \rm sum}}(a, b)$ between
$a$ and $b$ as the sum $K(b|a)+K(b|a)$
of the conditional complexities between the two items.
Therefore,
$D_{\mbox{\footnotesize \rm sum}}(a, b) = K(b|a)+K(a|b) \eqa
  K(R_b|S_a) +  K(R_a|S_b) $,
which can be substituted into (\ref{@}) to give:

\[
\log \G(a,b) \eqa -\frac{1}{2}D_{\mbox{\footnotesize \rm sum}}(a, b)
\]

or equivalently,
shifting to base $e$,

\begin{equation}\label{XX}
\ln \G(a,b) \eqa - \frac{\ln 2}{2} D_{\mbox{\footnotesize \rm sum}}(a, b) ,
\end{equation}
for almost all $a$ and $b$ (in the above sense of concentration
of $\Pr$-probability).

This means that $\G(a,b)$ is a negative exponential function of information
distance $D_{\mbox{\footnotesize \rm sum}}$, which is Shepard's Universal Law.
This is a surprising
result. It indicates that $\G(a,b)$, a measure of the confusability between
the items $a$ and $b$, has a specific functional relationship with a general
measure of distance, subject only to the mild assumption that the
probability distribution determining confusability is computable.

Two points concerning this result are worth noting. The first is that it
might appear that the result is somewhat {\it too} precise. Shepard's
Universal Law allows two free parameters, $A$ and $B$:

\[
\G(a,b) = Ae^{-B\cdot D(a, b)}
\]

whereas (\ref{XX}) has no apparent free parameters. But this disparity is
deceptive: The $\eqa$-symbol hides the parameter $A$, because it
gives equality---but only up to an additive constant term (which translates
into an multiplicative constant factor since (\ref{XX}) gives
the logarithmic version of the relation).
Moreover, the units for $D_{\mbox{\footnotesize \rm sum}}$
are arbitrary, because they depend on the
choice of a binary alphabet for measuring Kolmogorov complexity. Shifting
to an alphabet with a different number of elements (which can be viewed as
having any real value), values of $D_{\mbox{\footnotesize \rm sum}}$
will change by a multiplicative
constant, which can be interpreted  as parameter $B$.

Moreover, of course, our generalization of the universal law of generalization
doesn't hold for {\em all} items $a$ and $b$ but
for {\em almost all} items $a$ and $b$ (in the sense of concentration
of $\Pr $-probability).

The second point is that it might appear that the outcome of this result
provides some reason to prefer
$D_{\mbox{\footnotesize \rm sum}}$ over $D_{\max}$ as a preferred measure of
information distance in psychological contexts. But note that
they give the same values up to a multiplicative factor 2, since
we have noted above (\ref{*}) that
$D_{\max} \leq D_{\mbox{\footnotesize \rm sum}} \leq 2 D_{\max}$. But even
so, this apparent
preference between the two measures is merely a consequence of the specific
way in which Shepard defined $\G$.

\subsection{Improved Universal Law of Generalization}
It turns out that there are mathematical reasons to choose a slightly
different measure of the confusability between items $a, b$ than
initially chosen by Shepard. Define a new measure of
confusability as

\[
\G'(a,b)= \frac{\min\{ \Pr(R_b|S_a),\Pr(R_a|S_b) \}}
{\max \{\Pr(R_a|S_a), \Pr(R_b|S_b)\}},
\]
where we consider the ratio of (i) to (ii), such that
(i) is the minimum of the two probabilities that the stimulus
for $a$ elicits the response for $b$ or the stimulus for $b$
elicits the response for $a$, and (ii) is
the maximum of the two probabilities that the stimulus for $a$ elicits
the response for $a$ and the stimulus for $b$ elicits the response for $b$.
then analogous analysis to that above leads to a similar result.
Viz., from the earlier analysis argument we have
$-\log \Pr(R_b|S_a) \eqa K(b|a)$  (by noting $K(\Pr) \eqa 0$).
And moreover $K(b|a) \eqa 0$ for $b=a$ so that the precise form
of the denominator---whether $\min$, $\max$, square root
of product---doesn't matter since it will be a constant independent
of $a$ and $b$. The important part of the formula
is the nominator: note that the minimum for the conditional
probabilities in the formula translates into the maximum for
the related conditional Kolmogorov complexities. Thus, for almost
all $a, b$, in the sense of concentration of $\Pr$-probability, we obtain
$\log \G'(a,b) \eqa - D_{\max}(a, b)$ and therefore

\[
\ln \G'(a,b) \eqa - (\ln 2) D_{\max}(a, b)
\]
Straightforward substitution of the log-expressions of $\G$ and $\G'$
in the relation (\ref{*}) yields
$- \log \G'(a,b) \lea -2 \log \G (a,b) \lea -2 \log \G'(a,b)$.
That is, there are positive constants $C_1, C_2$ independent
of $a,b$ such that
 \[
\G'(a,b) \leq  C_1 \G(a,b) \leq C_2 \sqrt{\G'(a,b)}.
\]
for almost
all $a, b$, in the sense of concentration of $\Pr$-probability.
\footnote{Note that $\sqrt{\G'(a,b)} > \G'(a,b)$ since $0 < \G'(a,b) < 1$.
}
It seems likely that the two measures $\G'(a,b)$ and $\G(a,b)$ will be so
strongly positively correlated, in the empirical data, that the empirical
fits derived by Shepard for the Universal law using ``$\G$'' would be roughly
equally strong using ``$\G'$'', although we do not assess this directly.

There is a formal reason to prefer the $\G'(a,b)$-version as the proper
measure of confusability over the $\G(a,b)$ version, since it appeared
above that the negative logarithm of the $\G'(a,b)$ is precisely
(up to the $\eqa$ relation) the information distance $D_{\max}(a, b)$.
As we observed above, the latter has been shown in \cite{BGLVZ98} to
be the {\em universal} (that is optimal) cognitive distance. Viewing
a cognitive distance $D$ as defined in \cite{BGLVZ98}
as a code-length this means the following:
If we fix $b$ and let $a$ run over the possible items then define
the probability $P(a|b)$ of $a$ given $b$ by $P(a|b) = 2^{-D(a,b)}$.

It was shown in the cited reference that
$\sum_{a:a \neq b} 2^{-D(a,b)} \leq 1$ so that $P(a|b)$ is a proper
probability. In fact, the cognitive distance code of length $D(a,b)$,
the shortest binary program that serves
 to compute $a$ from $b$ and also to compute $b$ from $a$,
is length-equivalent to the Shannon-Fano code associated with
$P(a|b)$ and hence achieves the optimal (minimal)
expected code word length (the entropy
of $P$ by Shannon's Noiseless Coding Theorem, \cite{CT91})
among all prefix-codes.

Now lets go to the punch line: Since $D_{\max}(a, b)$ is the
minimal cognitive distance, minorizing all other cognitive distances,
up to a constant additive term,
its associated probability distribution
$P_{\G'} (a|b) := \G'(a,b) = 2^{-D_{\max}(a, b)}$, with $b$ fixed,
majorizes, up to a constant multiplicative factor,
{\em every} probability distribution $P_D (a|b)= 2^{-D(a,b)}$ with
$D(a,b)$ a cognitive distance.

That is, if we fix $b$ and consider
the probability of confusing any item $a$ with item $b$, according to
some semi-computable cognitive similarity criterion, as the negative
exponent of
the cognitive distance according to that similarity criterion,
then the confusion measure $\G'(a,b)$ is the largest such probability
incorporating confusability according to {\em all} semi-computable
(including all computable) cognitive
similarity criteria.

\section{Discussion}

We have shown that Shepard's Univeral Law of generalization follows, if we
assume that psychological distance is modelled as information distance. We
have also indicated that information distance is a highly general notion of
distance, which may be of broader psychological interest.

How does the derivation presented here relate to other formal work which
are described as providing derivations for the Universal Law of
generalisation, by Shepard \cite{Sh87} and Tenenbaum \& Griffiths
\cite{TenGriff}? These other derivation are, in fact, not directly related,
because these other derivations are concerned with a different and much
harder question: Why do items that are close in psychological distance tend
to have similar properties? This issue concerns the question of {\it
generalization}
proper--whereas the evidence that Shepard gathers concerns the
confusability between items.

Thus, we have here addressed the specific relationship evident in the data
that \cite{Sh87} encapsulates as the Universal Law. But an interesting
open question is whether the notion of information distance can be used to
address the question of generalization, as tackled by Shepard's and
Tenenbaum \& Griffith's results. Given the rich mathematical connections
between
the theory of Kolmogorov complexity and inductive inference and statistics
(e.g., Rissanen, \cite{Ri86,Ri89,Ri96}; Solomonoff, \cite{So64,So78};
Wallace
\cite{Wa68,Wa87}), it may be hoped some relationship between information
distance and generalization might be established.

Finally, we note that the generalization of the Universal Law that we have
outlined in this paper is attractive, because it applies in such a general
setting. Specifically, it does not presuppose that items correspond to
points in an internal multidimensional psychological space. This result
suggests a further line of empirical research, to determine whether the
Universal law does indeed hold in these more general circumstances. Such
research might investigate whether the Universal Law still holds, as we
would predict, even for stimuli, such as complex visual or linguistic
material, that seems unlikely to embed naturally into a multidimensional
psychological space. We hope that the present paper will serve as a
stimulus to empirical research of this kind.

\end{document}